\documentclass[sigconf]{acmart}
\usepackage{enumitem}
\usepackage[utf8]{inputenc}

\newcommand{\minisection}[1]{\vspace{0.04in} \noindent {\bf #1}\ \ }

\def\BibTeX{{\rm B\kern-.05em{\sc i\kern-.025em b}\kern-.08emT\kern-.1667em\lower.7ex\hbox{E}\kern-.125emX}}
    
\copyrightyear{2019}
\acmYear{2019}
\setcopyright{acmlicensed}

\acmConference[MM '19]{MM '19: ACM International Conference on Multimedia}{October 21--25, 2019}{Nice, France}
\acmBooktitle{MM '19: ACM International Conference on Multimedia, October 21--25, 2019, Nice, France}
\acmPrice{15.00}
\acmDOI{10.1145/1122445.1122456}
\acmISBN{978-1-4503-9999-9/18/06}
\begin{document}

\fancyhead{}
%
\title{SDIT: Scalable and Diverse Cross-domain Image Translation}

%
\author{Yaxing Wang, Abel Gonzalez-Garcia, Joost van de Weijer, Luis Herranz }
\email{{yaxing, agonzalez, joost, lherranz}@cvc.uab.es}
\orcid{1234-5678-9012}
\affiliation{%
  \institution{Computer Vision Center Universitat Aut\`{o}noma de Barcelona, Spain}
  \streetaddress{P.O. Box 1212}
  \postcode{43017-6221}
}



 





%
\begin{abstract}
  Recently, image-to-image translation research has witnessed remarkable progress. Although current approaches successfully generate diverse outputs or perform scalable image transfer, these properties have not been combined into a single method. To address this limitation, we propose SDIT: Scalable and Diverse image-to-image translation. These properties are combined into a single generator. The diversity is determined by a latent variable which is randomly sampled from a normal distribution.  The scalability is obtained by conditioning the network on the domain attributes. Additionally, we also exploit an attention mechanism that permits the generator to focus on the domain-specific attribute. We empirically demonstrate the performance of the proposed method on face mapping and other datasets beyond faces.
\end{abstract}



%

%
\maketitle

\section{Introduction}
\begin{figure}[t]
\centering
\includegraphics[width=1\columnwidth]{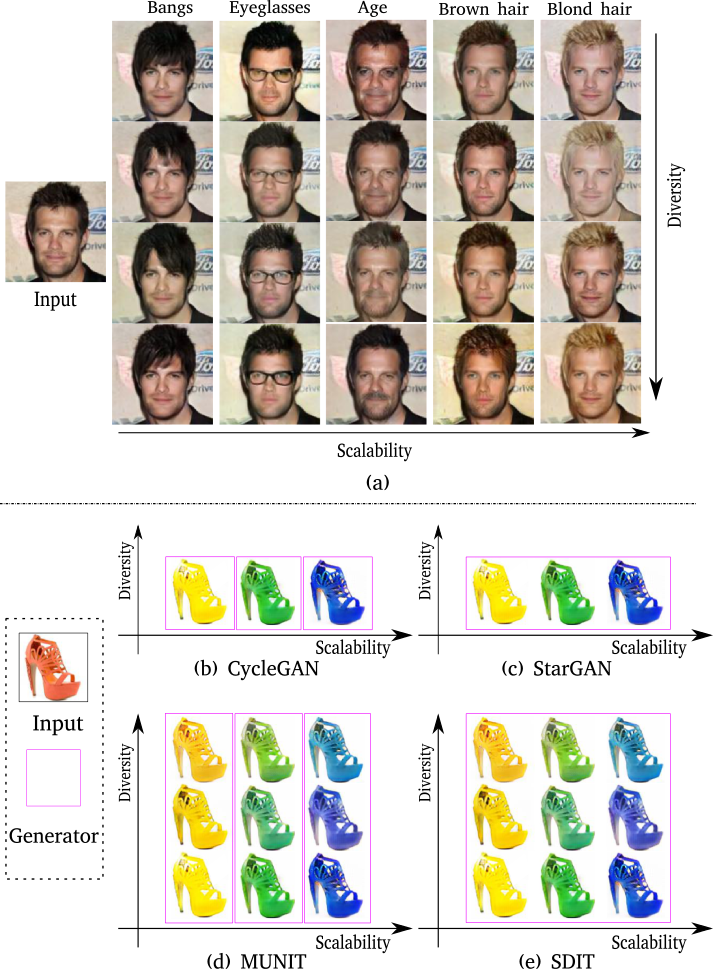}
\vspace{-4mm}
\caption{\small
\label{fig:introduction_framework}
(\textbf{a}) Example of diverse image translations for various attributes of our method generated by a single model. 
(\textbf{b-e}) Comparison to current unpaired image-to-image translation methods. Given four color subsets (\textit{orange}, \textit{yellow}, \textit{green}, \textit{blue}), the task is to translate images between the domains. (\textbf{b}) CycleGAN requires three independent generators (indicated by pink lines) which produce deterministic results. (\textbf{c}) StarGAN only requires a single generator but produces deterministic results. (\textbf{d}) MUNIT requires separate generators but is able to produce diverse results. (\textbf{e}) SDIT produces diverse results from a single generator.}
\vspace{-4mm}
\end{figure}

Image-to-image translation aims to build a model to map images from one domain to another. 
Many computer vision tasks can be interpreted as image-to-image translation, e.g.\ style transfer~\cite{gatys2016image}, image dehazing~\cite{dehaze_zhang_2018}, colorization~\cite{zhang2016colorful}, surface normal estimation~\cite{eigen2015predicting}, and semantic segmentation~\cite{long2015fully}. 
Face translation has always been of great interest in the context of image translation, and several methods~\cite{perarnau2016invertible, StarGAN2018, pumarola2018ganimation} have shown outstanding performance. 
Image-to-image translation can be formulated in a supervised manner when corresponding image pairs from both domains are provided, and unsupervised otherwise.
In this paper, we focus on unsupervised image-to-image translation with the two-fold goal of learning a model that has both scalability and diversity (see Figure~\ref{fig:introduction_framework}(a)).   

Recently, Isola \textit{et al.}~\cite{isola2016image} consider a conditional generative adversarial network to perform image mapping from input to output with paired training samples. One of the drawbacks, however, is that this method produces a deterministic output for a given input image. BicycleGAN~\cite{zhu2017toward} extended image-to-image translation to one-to-many mappings between images by training the model to reconstruct the noise used in the latent space, effectively forcing it to use it in the translations. To address the same concern, Gonzalez-Garcia \textit{et al}.~\cite{gonzalez2018image} explicitly exploit the feature representation, disentangling the latent feature into shared and exclusive representations, the latter being aligned with the input noise.

The above methods, however, need  paired images during the training process.   For many image-to-image translation cases, obtaining abundant annotated data remains very expensive or, in some cases, even impossible.  To relax the requirement of paired training images,  recent approaches have made efforts to address this issue. The cyclic consistency constraint~\cite{kim2017learning,yi2017dualgan,zhu2017unpaired} was initially proposed for unpaired image-to-image translation. Liu \textit{et al}.~\cite{liu2017unsupervised} assumes a shared joint latent distribution between the encoder and the decoder, then learns the unsupervised translation. 

Nonetheless, previous methods perform a deterministic one-to-one translation and lack diversity on its outputs, as shown in Figure~\ref{fig:introduction_framework}(b). For example,  given the task from orange (domain A) to yellow (domain B) the generator taking the orange shoes as input only synthesizes shows with a single shade of yellow.  Recently, the idea of non-deterministic outputs was extended to unpaired methods~\cite{huang2018multimodal,Lee2018drit} by disentangling the latent feature space into content and style and aligning the style code with a known distribution (typically Gaussian or uniform).
During inference, the model is able to generate diverse outputs by sampling different style codes from the distribution.  
The main drawback of these methods is that they lack scalability. As shown in Figure~\ref{fig:introduction_framework}(d) the orange shoes can be translated into many possible green shoes with varying green shades.
As the number of colors increases, however, the number of required domain-specific encoder-decoder pairs rises quadratically. 

IcGAN~\cite{perarnau2016invertible} initially performs face editing  by combining  cGAN~\cite{mirza2014conditional} with an attribute-independent encoder, and at the inference stage conducts face mapping for given face attributes. 
Recently, Yunjey \textit{et al.}~\cite{StarGAN2018} proposed StarGAN, a domain-independent encoder-decoder architecture for face translation that concatenates the domain label to the input image. Unlike the aforementioned non-scalable approaches~\cite{huang2018multimodal, Lee2018drit}, StarGAN is able to perform scalable image-to-image translation between multi-domains (Figure~\ref{fig:introduction_framework}(b)). StarGAN, however,  fails to synthesize diverse translation outputs.  

In this paper, we propose a compact and general architecture that allows for diversity and scalability in a single model, as shown in Figure~\ref{fig:introduction_framework}(e).  
Our motivation is that scalability and diversity are orthogonal properties that can be independently controlled. 
Scalability is obtained by using the domain label to train a single multi-domain image translator, preventing the need to train a encoder-decoder for each domain. 
Inspired by~\cite{dumoulin2017learned}, we employ Conditional Instance Normalization (CIN) layers in the generator to introduce the latent code and generate diverse outputs.
We explore the reasons behind CIN's success (Fig.~\ref{fig:CIN}) and discover the following limitation: CIN affects the entirety of the latent features and could possibly modify areas that do not correspond to the specific target domain.
To prevent this from happening, we include an attention mechanism that helps the model focus on domain-specific areas of the input image. 

Our contributions are as follows: 
\vspace{-3mm}
\setlist{nolistsep}
\begin{itemize}[noitemsep]
        \item We propose a compact and effective framework that combines both scalability and diversity in a single model. Note that current models only possess one of these desirable properties, whereas our model achieves both simultaneously.
        \item We empirically demonstrate the effectiveness of the attention technique for multi-domain image-to-image translation.
        \item We conduct extensive qualitative and quantitative experiments. The results show that our method is able to synthesize diverse outputs while being scalable to multiple domains.
\end{itemize}

\section{Related work}
\minisection{Generative adversarial networks.}Typical GANs~\cite{goodfellow2014generative} are  composed of two modules: a generator and a discriminator. The aim of the generator is to synthesize  images to fool the discriminator, while the discriminator distinguishes between fake images and real images. There  have been many variants of GANs~\cite{goodfellow2014generative} and  they show
remarkable performance on a wide variety of image-to-image translation tasks~\cite{isola2016image, zhu2017unpaired, yi2017dualgan, huang2018multimodal, Lee2018drit, pumarola2018ganimation}, super-resolution~\cite{ledig2017photo}, image compression~\cite{rippel2017real}, and conditional
image generation such as text to image\cite{zhang2017stackgan, zhang2017stackgan++, ma2018gan}, segmentation to image\cite{karras2017progressive,wang2018high} and domain adaptation~\cite{gong2012geodesic,ganin2015unsupervised,tsai2018learning, wu2018dcan,zhang2019synthetic, saito2017asymmetric,zou2018domain}.

\minisection{Conditional GANs.}Exploiting conditional image generation is an active topic in GAN research. 
Early methods considered incorporating into the model category information~\cite{mirza2014conditional,odena2016semi,odena2017conditional,StarGAN2018} or text description~\cite{reed2016generative, zhang2017stackgan, johnson2018image} for image synthesis. 
More recently, a wide variety of ideas have been proposed and used in several tasks such as image super-resolution~\cite{ledig2017photo}, video prediction~\cite{mathieu2015deep}, and photo editing~\cite{shu2017neural}. Similarly, we consider image-to-image translation conditioned on an input image and the label of the target domain.

\minisection{Image-to image-translation.}The goal of image-to-image translation is to learn a mapping between images of the source domain and images of the target domain. Given pairs of data samples, pix2pix~\cite{isola2016image} initially performed this mapping by using conditional GANs and relying on the real images.
This model, however, fails to conduct one-to-many mappings, namely, it cannot generate diverse outputs from a single input. BicycleGAN~\cite{zhu2017toward}  explicitly modeled the mapping between output and latent space, and aligned the latent distribution with a known distribution. 
Finally, the diverse outputs are performed by sampling from the latent distribution. 
Gonzalez-Garcia \textit{et al}.~\cite{gonzalez2018image}  disentangle the latent space into disjoint elements, which allows them to successfully perform cross-domain retrieval as well as one-to-many translation. 
Although these methods allow to synthesize diverse results, the requirement of paired data limits their application. 
Recently, the cycle consistency loss~\cite{kim2017learning,yi2017dualgan,zhu2017unpaired} is enforced into models to explicitly reconstruct the source sample, which is translated into the target domain and back, thus enabling translation using unpaired data.  
In addition, UNIT~\cite{liu2017unsupervised} aligns the latent space in two domains by assuming the similar domains share the same content. 
Although this approach shows remarkable results without paired data, they fail to perform diverse outputs. 
More recently, several image-to-image translation methods~\cite{almahairi2018augmented, choi2017stargan,pumarola2018ganimation, wang2018mix} enable  diverse results with the usage of noise or labels.

\minisection{Diversity of image-to-image translation.}Most recently, several approaches ~\cite{chen2016infogan, Lee2018drit, huang2018multimodal, gonzalez2018image, jakab2018unsupervised, yu2018singlegan, li2018twin}  consider to disentangle factors in feature space by enforcing a latent structure or regulating the structure distribution. 
Exploiting this disentangled representation enables the generator to synthesize diverse outputs by controlling style distribution. The key difference with the proposed method is that our method additionally performs \emph{scalable} image-to-image translation while still having diversity.

\minisection{Scalability of image-to-image translation.} The scalability aim is to conduct image-to-image translation across multiple domains by a single generator.  MMNet~\cite{wang2018mix} uses a shared encoder and a domain-independent decoder,  not only allowing to perform style learning but zero-pair image-to-image translation. Anoosheh \textit{et al.}~\cite{Anoosheh_2018} additionally consider encoder-decoder pairs  for each domain as well as the used techniques in CycleGAN~\cite{zhu2017unpaired}. IcGAN~\cite{perarnau2016invertible} and StarGAN~\cite{StarGAN2018} condition the domain label on the latent space and input, respectively.  
Our approach also works by imposing domain labels in a single generator, but simultaneously enabling the model to synthesize diverse outputs.

\minisection{Attention learning.}Attention mechanisms have been successfully employed for image-to-image translation. 
Current approaches~\cite{chen2018attention,mejjati2018unsupervised} learn an attention mask to enforce the translation to focus only on the objects of interest and preserve the background area.  GANimation~\cite{pumarola2018ganimation} uses \emph{action units} to choose regions from the input images that are relevant for facial animation. 
These methods exploit attention mechanisms at the image level.  
Our method, on the other hand, learns feature-wise attention maps, which enables us to control which features are modified during translation. 
Therefore, our attention maps are highly effective at restricting the translation to change only domain-specific areas (e.g.\ forehead region when modifying the `bangs' attribute).

\begin{figure*}
    \centering
   \includegraphics[width=\textwidth]{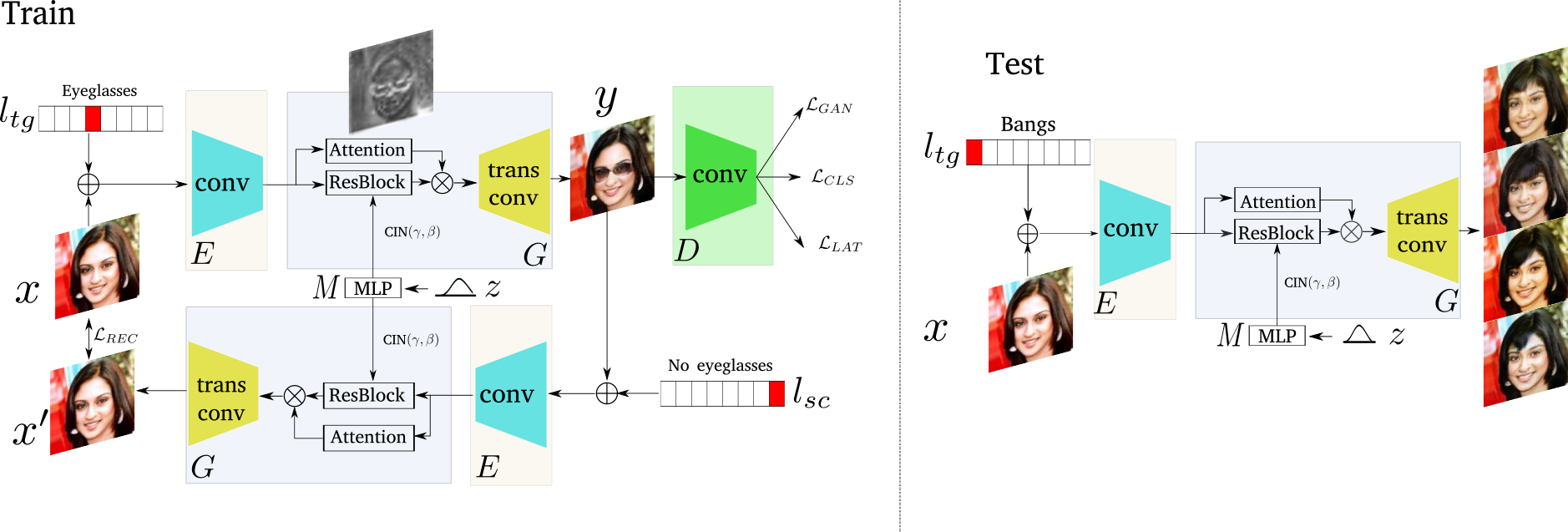}\vspace{-4mm}
   \caption{\small \textbf{Model architecture}. (\textit{Left}) The proposed approach is composed of two main parts: a discriminator $D$ to distinguish the generated images and the real images; and the set of the encoder $E$, multilayer perceptron $M$ and the generator $G$, containing the attention block, residual blocks with CIN, and the transposed convolutional layers. (\textit{Right}) At test time, we can generate multiple plausible translations in the desired domain using a single model.}\vspace{-3mm}
   \label{fig:basic_framework}
\end{figure*}

\section{Scalable and Diverse Image Translation}

Our method must be able to perform multi-domain image-to-image translation. 
We aim to learn a model with both scalability and diversity. By scalability we refer to the property that a single model can be used to perform translations between multiple domains. By diversity we refer to the property that given a single input image, we can obtain multiple plausible output translations by sampling from a random variable. 

\subsection{Method Overview} 
Here we consider two domains: source domain $\mathcal{X} \subset \mathbb{R}^{H\times W\times 3} $ and target domain $\mathcal{Y} \subset \mathbb{R}^{H\times W\times 3} $ (it can trivially be extended to multiple domains). As illustrated in Figure~\ref{fig:basic_framework}, our framework is composed of four neural networks: encoder $E$, generator $G$, multilayer perceptron $M$, and discriminator $D$. 
Let $x  \in \mathcal{X}$ be the input source image and $y  \in \mathcal{Y}$ the target output, with corresponding labels  $l_{sc}\in\left\{1,\ldots,C\right\}$ for the source and $l_{tg}\in\left\{1,\ldots,C\right\}$ for the target. 
In addition,  let $z \in \mathbb{R}^{Z}$ be the latent code, which is sampled from a Gaussian distribution.

An overview of our method is provided in Figure~\ref{fig:basic_framework}. To address the problem of scalability we introduce the target domain as a conditioning label to the encoder, $E(x, l_{tg})$. The diversity is introduced by the latent variable $z$, which is mapped to the input parameters of a Conditional Instance Normalization (CIN) layer~\cite{dumoulin2017learned} by means of the multilayer perceptron $M(z)$. The CIN learns an additive ($\beta$) and a multiplicative term ($\gamma$) for each feature layer. Both the output of the encoder $E$ and the multilayer perceptron $M$ are used as input to the generator $ G(E(x, l_{tg}), M(z))$. The generator $G$ outputs a sample $y$ of the target domain. Sampling different $z$ results into different output results $y$. The unpaired domain translation is enforced by a cycle consistency~\cite{kim2017learning,yi2017dualgan,zhu2017unpaired}: taking as input the output $y$ and the source category $l_{sc}$, we reconstruct the input image $x$ as $G(E(G(E(x, l_{tg}), M(z)), l_{sc}), M(z))$.  The encoder $E$, the multilayer perceptron $M$, and the generator $G$ are all shared. 

The function of the discriminator $D$ is threefold.  It produces three outputs: $x\rightarrow \left \{ D_{src}\left ( x \right ), D_{cls}\left ( x \right ), F_{rec}\left ( x \right ) \right \}$. Both  
$  D_{src}\left ( x \right ) $ and $ D_{cls}\left ( x \right ) $ represent probability distributions, while  $F_{rec}\left ( x \right ) $ is a regressed code.  The goal of  ${ D_{src}\left ( x \right )}$ is to distinguish between real samples and generated images in the target domain. The auxiliary classifier ${ D_{cls}\left ( x \right )}$  predicts the target label and allows the generator to perform domain-specific output conditioned on it. 
This was found to improve the quality of the conditional GAN~\cite{odena2017conditional}. 
Similarly to previous methods~\cite{chen2016infogan,huang2018multimodal} we reconstruct the latent input code in the output ${ F_{rec}\left ( x \right )}$. This was found to lead to improved diversity. 
Note that $F_{rec}$ is just used for generated samples, as $F_{rec}$ aims to reconstruct the latent code, which is not defined for real images. 

We shortly summarize here the differences of our method with respect to the most similar approaches. 
StarGAN~\cite{StarGAN2018} can also generate outputs on multiple domains, but: (1) it learns a scalable but deterministic model, while our method additionally  obtains diversity via the latent code; 
(2) we explicitly exploit an attention mechanism to focus the generator on the object of interest. 
Comparing against both MUNIT~\cite{huang2018multimodal} and DRIT~\cite{Lee2018drit}, which perform diverse image-to-image translation but without being scalable, our method: (1) employs  the domain label to control the  target domain, allowing to conduct image-to-image translation among multiple domains with a single generator; (2) avoids the need for domain-specific style encoders,  effectively saving computational resources; (3) considers attention to avoid  undesirable changes in the translation; and (4) experimentally proves that the bias of CIN is the key factor to make the generator achieve the diversity, whereas the multiplicative term was only found to play a minor role.

\subsection{Training Losses} 
The full loss function consists of several losses:  the \textit{adversarial loss} that discriminates the distribution of synthesized data and the real distribution in target domain, \textit{domain classification loss} which contributes to the model $\{E$, $G\}$ to learn the specific attribute for a given target label, the \textit{latent code reconstruction loss} regularizes the latent code to improve diversity and avoids the problem of partial mode collapse, and the \textit{image reconstruction loss} that guarantees that the translated image keeps the structure of the input images. 

\minisection{Adversarial loss.} We employ GANs~\cite{goodfellow2014generative} to distinguish the generated images from the real images 
\begin{equation}
\begin{aligned}
& \mathcal{L}_{GAN} = \mathbb{E}_{x\sim \mathcal{X}}\left[ \log D_{src}(x) \right] \\ & +  \mathbb{E}_{x\sim \mathcal{X}, z\sim p(z)}\left[ \log (1 - D_{src}(G(E(x, l_{tg}), M(z)))) \right],
\end{aligned}
\end{equation} 
where the discriminator tries to differentiate between generated images from the generator and real images,  while $G$ tries to fool the discriminator taking the output of $M$ and the output of $E$ as input. The final loss function is optimized by the minimax game  
\begin{equation}
\left \{ E^{\ast }, G^{\ast }, D^{\ast } \right \}= \arg\min_{E, G}\max_{D}\mathcal{L}_{GAN}.
\end{equation} 

\minisection{Domain classification loss.}
In this paper, we consider Auxiliary Classifier GANs (AC-GAN)~\cite{odena2017conditional} to control domains. The discriminator aims to output a probability distribution over given input images $y$ and domain label, in consequence $E$ and $G$ synthesize the domain-specific images. 
We share the  discriminator model except for the last layer and optimize the triplet $\left\{E,G,D\right\}$  by the cross-entropy loss. The final domain classification loss for generated samples, real samples, and total are
\begin{equation}
\resizebox{0.92\hsize}{!}{
$\mathcal{L}_{FAKE}\left(E, G\right)= -\mathbb{E}_{x\sim \mathcal{X}, z\sim p(z)}\left[\log\left(D_{cls}\left(l_{tg}|G(E(x, l_{tg}),M(z))\right)\right)\right]$,
}\label{eq: l_fake}
\end{equation}
\begin{equation}
\mathcal{L}_{REAL}\left(D\right)=-
\mathbb{E}_{x\sim \mathcal{X}}\left[\log\left(D_{cls}\left(l_{sc}|x\right)\right)\right],
\label{eq: l_real}
\end{equation}
\begin{equation}
\mathcal{L}_{CLS}=\mathcal{L}_{REAL} + \mathcal{L}_{FAKE},
\label{eq: l_cls}
\end{equation}
respectively.  Given domain labels $l_{sc}$ and $l_{tg}$ these objectives are able to minimize the classification loss so that the model explicitly generates domain-specific outputs.  

\minisection{Latent code reconstruction loss.} The lack of constraints on the latent code results in the generated images suffering from partial mode collapse as the latent code is ignored. We use the discriminator to predict the latent code, which forces the network to use it for generation:  
\begin{equation}
\mathcal{L}_{LAT}\left(E, G, D\right)= \mathbb{E}_{x\sim \mathcal{X}, z\sim p(z)}\left [ \left \|  F_{rec}(x) - z \right \|_{1}
\label{eq: latent_code} \right ]
\end{equation}
\minisection{Image reconstruction loss.} Both adversarial loss and classification loss fail to keep the structure of the input. 
To avoid this, we formulate the image reconstruction loss as

\vspace{-3pt}
\begin{equation}
\begin{aligned}
& y=G\left ( E\left ( x, l_{tg} \right ), M(z) \right ), \\
&{x}'=G\left ( E\left (y, l_{sc} \right ), M(z) \right ), \\ &\mathcal{L}_{REC}=\mathbb{E}_{x\sim \mathcal{X},{x}' \sim \mathcal{X^{'}}}\left [\left \| x - {x}' \right \|_{1} \right ].
\end{aligned}
\end{equation}

\minisection{Full Objective.}
The full objective function of our model is:
\vspace{-2pt}
\begin{equation}
\label{eq:loss}
\begin{aligned}
& \underset{E,G}{\min} \ \underset{D}{\max} \  \lambda_{GAN}\mathcal{L}_{GAN} +  \lambda_{FAKE}\mathcal{L}_{FAKE} \\
&+\lambda_{REAL}\mathcal{L}_{REAL}+  \lambda_{LAT}\mathcal{L}_{LAT} + 
\lambda_{REC}\mathcal{L}_{REC}
\end{aligned}
\end{equation}
where $  \lambda_{GAN}$,   $\lambda_{FAKE}$,  $\lambda_{REAL}$,   $\lambda_{LAT}$, $\lambda_{REC}$ are hyper-parameters that balance the importance of each iterm.

\subsection{Attention-guided generator}

The attention mechanism encourages the generator to locate the domain-specific area relevant to the target domain label. Let $e =  E\left ( x, l_{tg} \right )$ be the output of the encoder. We propose to localize the CIN operation by introducing an attention mechanism. Only part of the encoder output $e$ should be changed to obtain the desired diversity. We separate the signal $e$ into two parallel residual blocks $T^c$ and $T^a$.
The CIN is applied to the residual block according to $f =  T^{c}\left ( e, M\left ( z \right ) \right ) $. We estimate the attention with a separate residual block according to $a = T^{a}\left (e \right )$.  We then combine the original encoder output and the CIN output using attention:
\vspace{-4pt}
\begin{equation}
h = (1 - a)\cdot e + a\cdot f.
\label{eq: attention_full} 
\end{equation}
In~\cite{pumarola2018ganimation}, an \textit{attention loss} regularizes the attention maps, since they quickly saturate to 1. 
In contrast, we  employ the attention in the bottleneck features, and experimentally prove that the attention masks can be easily learned. 
This makes the task easier due to lower resolution in the bottleneck, and avoids the need to tune the attention hyperparameter.
Finally, our attention mechanism does not add any new terms to the overall optimization loss in~\eqref{eq:loss}.
 
\section{Experimental setup}

\minisection{Training setting.}Our model is composed of four sub-networks: encoder $E$, multilayer perceptron $M$, generator $G$, and discriminator $D$. The encoder contains 3 convolutional layers and 6 blocks. Each convolutional layer uses $4 \times 4$ filters with stride 2, except for the first one which uses $7 \times 7$ with stride 1, and each block contains two convolutional layers with $3 \times 3$ filters and stride of 1. $M$ consists of two fully connected layers with 256 and 4096 units. 
The generator $G$ comprises  ResBlock layers, attention layers and two fractionally strided convolutional layers. 
The ResBlock consists of 6 residual blocks, as in the encoder $E$, but including CIN layers. 
The CIN layers take the output of $E$ and the ouput of the $M$ as input. Except for six blocks like the CIN layers, the attention layers also use additional convolutional layers with sigmoid activations on top. 
For the discriminator $D$, we use six convolutional layers with $4 \times 4$ and stride 2, followed by three parallel sub-networks, each of them containing one convolutional layer with $3 \times 3$ filters and stride 1, except for the branch to output $F_{rec}$ which uses an additional fully connected layer from 32 units to 8.
Note how $M$ adds around 1M parameters to the architecture.
    
All models are implemented in PyTorch~\cite{paszke2017automatic} and released\footnote{The codes are available at \url{https://github.com/yaxingwang/SDIT} }. We randomly initialize the weights following a Gaussian distribution, and optimize the model using Adam~\cite{kingma2014adam} with batch size 16 and 4 for face and non-face datasets, respectively. The learning rate is 0.0001, followed the exponential decay rates $\left ( \beta_{1},   \beta_{2} \right ) = \left ( 0.5, 0.999 \right )$. In all experiments, we use the following hyper-parameters:  $  \lambda_{GAN} =10 $,   $\lambda_{FAKE} = 1$,  $\lambda_{REAL} = 1$,   $\lambda_{LAT} = 10$ and $\lambda_{REC} = 800$. We use Gaussian noise to the latent code with zero mean and a standard deviation of 1.

\subsection{Datasets}We consider several datasets to evaluate our models. In order to verify the generality of our method, the datasets were chosen to cover a variety of cases, including  faces (CelebA), object (Color), and scenes (Artworks). 

\textbf{CelebA~\cite{liu2015faceattributes}.} The Celeb Faces Attributes is a face dataset of celebrities with 202,599 images and 40 attribute labels per face. To explicitly preserve the face ratio, we crop the face size of $178 \times 218 $ and resize it to $128 \times 128$.
We leave out 2000 random images for test and train with the rest.  

\textbf{Color dataset~\cite{yu2018weakly}.} We use the dataset collected by Yu \textit{et.al}~\cite{yu2018weakly}, which consists of 11 color labels, each category containing 1000 images. In order to easily compare to the non-scalable baselines which need train one independent model for each domain pair, we use only four colors (\textit{green}, \textit{yellow}, \textit{blue}, \textit{orange}). We resize all images to $128 \times 128$.  We collected 3200 images for the train set and 800 images for the test set. 

\textbf{Artworks~\cite{zhu2017unpaired}.} We also illustrate SDIT in an artwork setting~\cite{zhu2017unpaired}. This includes real images (\textit{photo}) and three artistic styles (\textit{Monet}, \textit{Ukiyo-e}, and \textit{Cezanne}). The training set contains 3000 (photo), 700 (Ukiyo-e), 500 (Cezanne) and 1000 (Monet) images, while  the test set are: 300 (photo), 100 (Ukiyo-e), 100 (Cezanne) and 200 (Monet) images. All image are resized to $256 \times 256$.

\subsection{Evaluation Metrics}To validate our approach, we consider the three following metrics. 

\textbf{LPIPS.} In this paper, LPIPS~\cite{zhang2018perceptual} is used to  compute the  similarity of pairs of images from the same attribute. LPIPS takes larger values if the generator has more diversity. 
In our setting, we generate 10 samples given an input image via different random codes.

\textbf{ID distance.} The key point of face mapping is to  preserve the \textit{identity} of the input, since an identity change is unacceptable for this task. 
To measure whether two images depict the same identity, we consider \textit{ID distance}~\cite{wang2019controlling}, which represents the difference in identity between pairs of input and translated faces.
More concretely, given a pair of input and output faces, we extract the identity features represented by the VGGFace~\cite{Parkhi15} network, and compute the distance between these features.
VGGFace is trained on a large face dataset and is robust to appearance changes (e.g. illumination, age, expression, etc.). 
Therefore, two images of the same person should have a very small value. 
We only use this evaluation metric for CelebA.
We use all 2000 test images as input and generate 10 output images, which in total amounts to 20,000 pairs.

\begin{figure}[t]
\centering
\includegraphics[width=1\columnwidth]{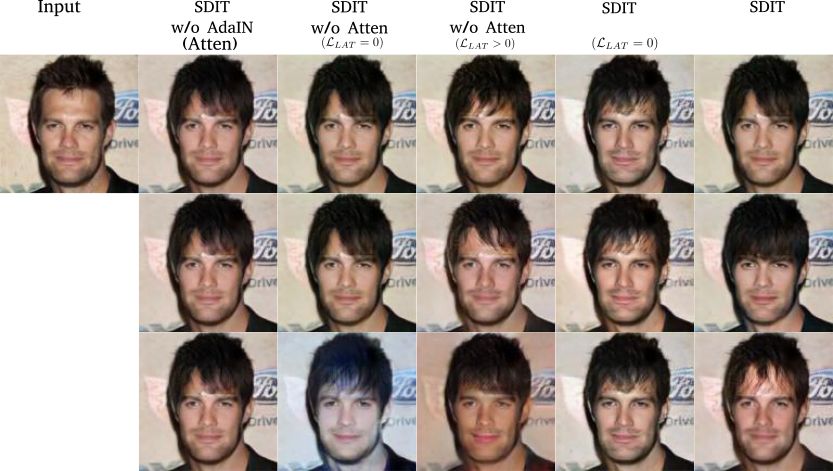}
\caption{\small \label{fig:ablation_study_losses} Ablation study of different variants of our method. We show results for the face task of adding `bangs'. We display three random outputs for each variant of the method.}
\vspace{-3mm}
\end{figure}
\begin{table}[t]
\setlength{\tabcolsep}{10.5pt}
\centering

\resizebox{1\columnwidth}{!}{
\begin{tabular}{c|ccc|c|c}
\hline
Method & Atten & CIN & $\mathcal{L}_{LAT}$  &ID Distance & LPIPS  \\
\hline
SDIT w/o CIN (Atten)  & Y & N & N & 0.061  &0.408    \\
SDIT w/o Atten ($\mathcal{L}_{LAT} = 0 $) & N & Y& N  & 0.063  &  0.409      \\
SDIT w/o Atten ($\mathcal{L}_{LAT} > 0 $) & N & Y & Y  & 0.070    & \textbf{0.432}      \\
SDIT  ($\mathcal{L}_{LAT} = 0 $) & Y & Y & N  & 0.063  &0.412         \\
SDIT  & Y & Y & Y &  \textbf{0.060} & 0.424          \\

\hline
\end{tabular}
}
 \caption{\small ID distance (lower, better) / LPIPS  (higher, better) for different variants of our method. Atten: attention, Y: yes, N: no.}
  \vspace{-10mm}
  \label{table:ablation_study}%
\end{table}
\begin{figure*}[t]
    \centering
   \includegraphics[width=\textwidth]{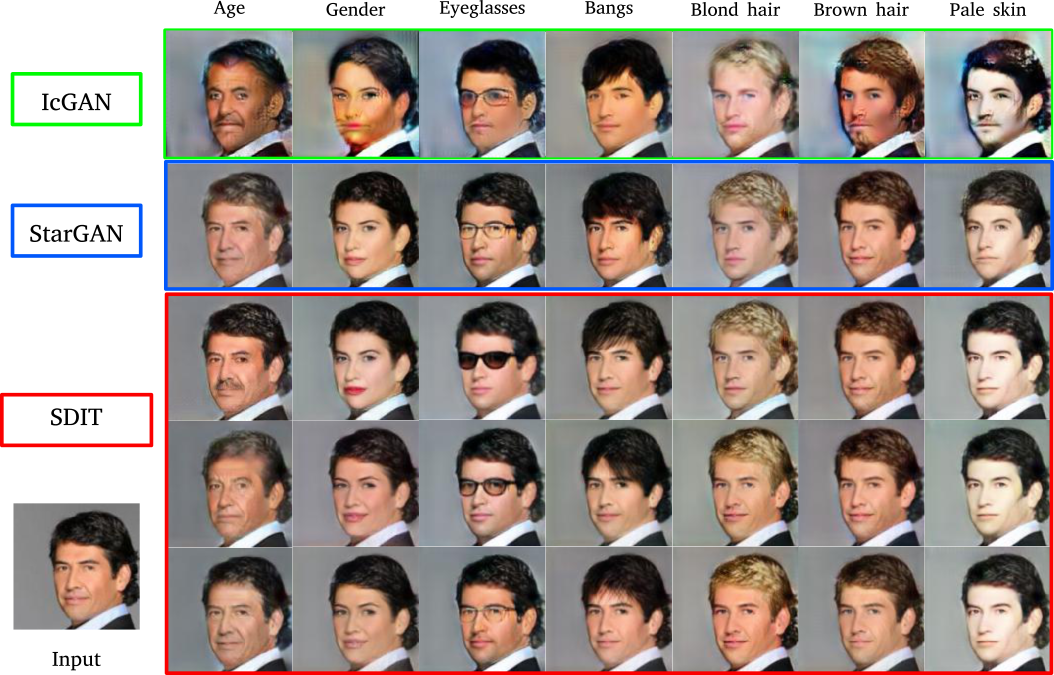}
   \caption{ \small Qualitative comparison to the baselines. The input face image is at the left bottom and the remaining columns show the  attribute-specific mapped images. The first two lines show the translated results of the IcGAN~\cite{perarnau2016invertible} and StarGAN~\cite{StarGAN2018}, respectively, while the remaining rows are from the proposed method.  }\vspace{-2mm}
   \label{fig:face_compara_baselines}
\end{figure*}

\textbf{Reverse classification.} One of the methods to evaluate conditional image-to-image translation is to train a reference classifier on real images and test it on generated images~\cite{wang2018transferring, wu2018memory}. The reference classifier, however, fails to evaluate diversity, since it may still report a high accuracy even when the generator encounters mode-collapse for a specific domain, as shown on the third column of Figure~\ref{fig:ablation_study_losses}.  Following~\cite{shmelkov2018good,wu2018memory}, we use the \textit{reverse classifier} which is trained using translated images for each target domain and evaluated on real images for which we know the label. 
Lower classification errors indicate more realistic and diverse translated images.  

\section{Experimental Results}
In Section~\ref{sec:baselines} we introduce several baselines against which we compare our model, as well as multiple variants of our model. Next, we evaluate the model on faces in Section~\ref{sec:face}. Finally, in Section~\ref{sec:color} and Section~\ref{sec:scene}, we analyze the generality of the model to color translation and scene translation.

\subsection{Baselines and variants}\label{sec:baselines}

\begin{table*}
\setlength{\tabcolsep}{2.5pt}
\centering

\resizebox{1\textwidth}{!}{
\begin{tabular}{c|cccccccccc|c}
\hline
Method  & \rotatebox{0}{Bangs}  
&\rotatebox{0}{Age\;}  
&\rotatebox{0}{Gender\;}  
&\rotatebox{0}{Smiling\;} 
&\rotatebox{0}{Wearing hat\;}
&\rotatebox{0}{Pale skin\;}  
&\rotatebox{0}{Brown hair\;} 
&\rotatebox{0}{Blond hair\;} &\rotatebox{0}{Eyeglasses\;} 
&\rotatebox{0}{Mouth open\;}  
&\rotatebox{0}{Mean\;}  
\\
\hline
StarGAN~\cite{StarGAN2018}
 &\textbf{0.067}/0.427 &\textbf{0.065}/0.428 &\textbf{0.068}/0.428 &\textbf{0.061}/0.427 &\textbf{0.075}/0.427 &\textbf{0.064}/0.421 &0.060/0.418 &0.067/0.426 &\textbf{0.066}/0.435 &0.059/0.429 &\textbf{0.065}/0.427  \\
 \hline
IcGAN~\cite{perarnau2016invertible}
  &{0.118/0.430}   &{0.097/0.431}  &{0.094/0.430}    &{0.121/0.430}    &{0.102/0.429}    &{0.10/0.430}   &{0.127/0.424}  &{0.113/0.421}    &{0.097/0.425}     &{0.116/0.438}    &{0.108/0.432}  \\   
 \hline
SDIT
 &0.068/\textbf{0.456} &\textbf{0.065}/\textbf{0.447} &0.069/\textbf{0.444} &\textbf{0.061}/\textbf{0.449} &0.076/\textbf{0.458} &0.065/\textbf{0.439} &\textbf{0.058}/\textbf{0.443} &\textbf{0.067}/\textbf{0.442} &\textbf{0.066}/\textbf{0.458} &\textbf{0.058}/\textbf{0.457} &\textbf{0.065}/\textbf{0.451}  
\\ 
\hline
Real data
 &-/0.486 &-/0.483 &-/0.484 &-/0.480 &-/0.489 &-/0.479 &-/0.492 &-/0.490 &-/0.492 &-/0.489 &-/0.486  
\\
\hline

\end{tabular}
}
\caption{\small ID distance (lower, better) / LPIPS (higher, better) on CelebA dataset.}
\vspace{-8mm}
\label{table:compara_to_baseline}
\end{table*}

We compare our method with the following baselines.
For all baselines, we use the authors' original implementations and recommended hyperparameters.
We also consider different configurations of our proposed SDIT approach. 
In particular, we study variants with and without CIN, attention, and latent code reconstruction.

\textbf{CycleGAN~\cite{zhu2017unpaired}.} CycleGAN is composed of two pairs of domain-specific encoders and decoders.  The full objective is optimized with an adversarial loss and a cycle consistency loss.  

\textbf{MUNIT~\cite{huang2018multimodal}.} MUNIT disentangles the latent distribution into the content space which is shared between two domains,  and the style space which is domain-specific and aligned with a Gaussian distribution.  
At test time, MUNIT takes as input the source image and different style codes to achieve diverse outputs.

\textbf{IcGAN~\cite{perarnau2016invertible}.} IcGAN explicitly maps the input face into a latent feature, followed by a decoder which is conditioned on the latent feature and a target face attribute. In addition, the face attribute can be explicitly reconstructed by an inverse encoder.  

\textbf{StarGAN~\cite{StarGAN2018}.} StarGAN shares the  encoders and decoders for all domains. The full model is trained by optimizing the adversarial loss, the reconstruction loss and the cross-entropy loss, which controls that the input image is translated into a target image.

\subsection{Face translation}\label{sec:face}

We firstly conduct an experiment on the CelebA~\cite{liu2015faceattributes} dataset to compare against ablations of our full model. Next, we compare SDIT to the baselines. For this case, we consider IcGAN and StarGAN, both of which show outstanding results for face synthesis.

\minisection{Ablation study.} We performed an ablation study comparing several variants of SDIT in terms of model diversity. We consider five attributes, namely \textit{bangs}, \textit{blond hair}, \textit{brown hair}, \textit{young}, and \textit{male}. Figure~\ref{fig:ablation_study_losses} shows the translated images obtained with different variants of our method. As expected, SDIT with only \textit{attention} (second column of Figure~\ref{fig:ablation_study_losses}) fails to synthesize diverse outputs, since the model lacks the additional factors (e.g.\ noise) to control this. Both the third and fourth columns show that adding CIN to our method without attention generates diverse images. 
Their quality, however, is unsatisfactory and the model suffers from  partial mode collapse, since CIN operates on the entire image, rather than being localized by the attention mechanism to the desired area (e.g.\ the bangs). 
Combining both CIN and attention but without the latent code reconstruction ($\mathcal{L}_{LAT} = 0 $) leads to little diversity, as shown in the fifth column. 
Finally, our full model (last column) achieves the best results in terms of quality and diversity.

For quantitative evaluation, we report the results in terms of the ID distance and LPIPS. As shown in Table~\ref{table:ablation_study}, the SDIT models without CIN or $\mathcal{L}_{LAT}$ generate less diverse outputs according to LPIPS scores.  Using $\mathcal{L}_{LAT}$  without attention contributes to improve the diversity. It has a higher LPIPS, but this could be because it is adding unwanted diversity (e.g.\ the red lips in the fourth column of Figure~\ref{fig:ablation_study_losses}). This may explain its higher ID distance. 
Combining both attention and $\mathcal{L}_{LAT} > 0$ (i.e.\ the full SDIT model) encourages the results to have better targeted diversity, as reported in the last row of Table~\ref{table:ablation_study}. The preservation of identity is crucial for the facial attribute transfer task, and thus we keep both attention and the reconstruction loss in the following sections.     

\begin{figure}[t]
\centering
\includegraphics[width=1\columnwidth]{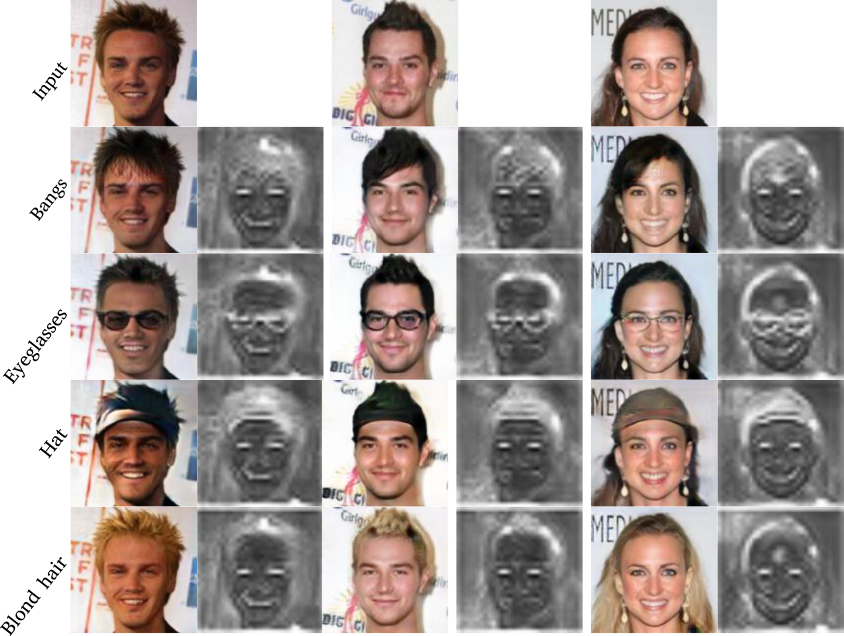}

\caption{\small \label{fig:attention} Generated images and learned attention maps for three input images. 
For each of them we present multi-domain outputs and attribute-specific attention. } 
\vspace{-3mm}
\end{figure}

\begin{figure}[t]
\centering
\includegraphics[width=0.9\columnwidth]{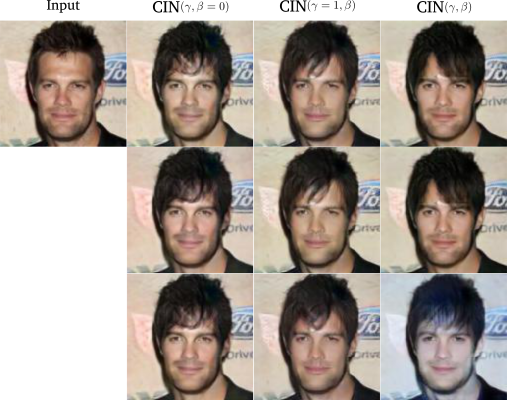}
\caption{\label{fig:CIN} \small Ablation study on CIN.  We compare three cases: $(\gamma, \beta = 0)$, where $\gamma$ is learnable; $(\gamma =1 , \beta)$, where $\beta$ is learnable; and $(\gamma, \beta)$, where both $\gamma$ and  $\beta$ are learnable.}
\vspace{-3mm}
\end{figure}

\minisection{Attention.} Figure~\ref{fig:attention} shows the attention maps for several translations from the face dataset. We note that our method explicitly learns the attribute-specific attention for a given face image (e.g.\ \textit{eyeglasses}), and generates the corresponding outputs. In this way, attention enables to modify only attribute-specific areas of the input image. 
This is a key factor to restrict the effect of the CIN, which otherwise would globally process the entire feature representation.

\minisection{CIN learning.} We explain here how CIN contributes to the diversity of the generator. In this experiment, we only consider CIN without attention nor latent code reconstruction. 
The operation performed by CIN on a feature $e$ is given by:
\begin{equation}
\textup{CIN}(e; z) = \gamma(z)  \left( \frac{e -  \mu(e)}{\delta (e)}\right)+\beta (z)
\label{eq: attention_3} 
\end{equation}
where $e$ and $z$ are the output of encoder $E$ and latent code $z$, respectively;  $\gamma, \beta$ are affine parameters learned from $M$ and $\mu(e)$, $\delta(e)$ are the mean and standard deviation. As shown in the second column of Figure~\ref{fig:CIN}, only learning $\gamma$ fails to output diverse images, while only learning $\beta$ already generates diverse results (third column of Figure~\ref{fig:CIN}), clearly indicating that $\beta$ is the key factor to diversity. Updating the two parameters obtains a similar performance in this task.
However, $\beta$ could be ignored by the network. Therefore we introduced the latent code reconstruction loss, Eq.~\ref{eq: latent_code}, which helps to avoid this.   

\minisection{Comparison against baselines.} 
Figure~\ref{fig:face_compara_baselines} shows the comparison to the baselines on test data. 
We consider ten attributes: \textit{bangs}, \textit{blond hair}, \textit{brown hair}, \textit{young}, \textit{male}, \textit{mouth slightly open}, \textit{smiling}, \textit{pale skin}, \textit{wearing hat}, and \textit{eyeglasses}.  Although both IcGAN and StarGAN are able to perform image-to-image translation to each domain, they fail to synthesize diverse outputs. Moreover, the performance of IcGAN is unsatisfactory and it fails to keep the personal identity.  Our method not only enables the generation of realistic and diverse outputs, but also allows scalable image-to-image translation. 
Note that both StarGAN and our method use a single model. The visualization shows that scalability and diversity can be successfully integrated in a single model without conflict. Taking adding \emph{bangs} as an example translation; the generated bangs with different directions do not impact the classification performance or the adversarial learning, in fact possibly contribute to the adversarial loss, since the CIN layer slightly reduces the compactness of the network, which increases the freedom of the generator.

As we can see in Table~\ref{table:compara_to_baseline}, our method obtains the best scores in both LPIPS and ID distance. In the case of LPIPS, the mean value of our method is 0.451, while IcGAN and StarGAN achieve 0.432 and 0.427 respectively. 
This clearly indicates that SDIT can successfully generate multimodal outputs using a single model. Moreover, the low ID distance indicates that SDIT effectively preserves the identity, achieving a competitive performance with StarGAN. 
Note that here we do not compare to CycleGAN and MUNIT because these methods require a single generator to be trained for each pair of domains. 
This is unfeasible for this task, because each attribute combination would require a different generator.

\begin{figure*}[t]
    \centering
   \includegraphics[width=\textwidth]{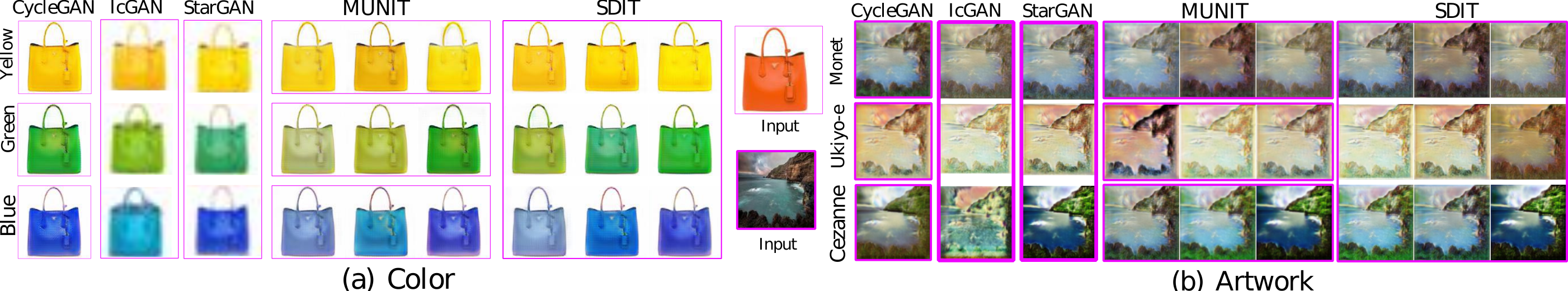}\vspace{-3mm}
   \caption{\small Examples of scalable and diverse inference of multi-domain translations on (a) color dataset and (b) artworks dataset. In both cases, the first column is the input, the next three show results for CycleGAN~\cite{zhu2017unpaired}, IcGAN~\cite{perarnau2016invertible}, and StarGAN~\cite{StarGAN2018}, respectively, followed by three samples from MUNIT~\cite{huang2018multimodal} in next three columns and three samples from SDIT in the last three. Each row indicates a different domain.}
   \vspace{-2mm}
   \label{fig:color_art}
\end{figure*}

\subsection{Object translation}\label{sec:color}

The experiments in the previous section were conducted on a face dataset, in which all images have a relatively similar content and structure (a face on a background).
Here we consider the color object dataset to show that SDIT can be applied to datasets that lack a common structure. This dataset contains a wide range of different objects which greatly vary in shape, scale, and complexity. This makes the translation task more challenging.

\minisection{Qualitative results.}
Figure~\ref{fig:color_art}(a) compares image-to-image translations obtained with CycleGAN~\cite{zhu2017unpaired}, IcGAN~\cite{perarnau2016invertible}, StarGAN~\cite{StarGAN2018}, MUNIT~\cite{huang2018multimodal} and the proposed method.  We can see how SDIT clearly generates highly realistic and attribute-specific bags with different color shades, which is comparable to the results of MUNIT. Other baselines,  however,  only generate one color shade. The main advantage of SDIT is the \textit{scalability}, as SDIT explicitly synthesizes the target color image (\textit{yellow}, \textit{green}, or \textit{blue}) using a single generator.

\minisection{Quantitative results.} The qualitative observations above are validated here by quantitative evaluations. Table~\ref{table:color} compares the results of SDIT to the baseline methods. Our method outperforms both baseline methods on LPIPS despite only using a single model. For the classification accuracy, CycleGAN, IcGAN and StarGAN produce a lower score, since it is not able to generate diverse outputs for a given test samples. Both MUNIT and SDIT have a similar performance. However, for both CycleGAN and MUNIT training all pairwise translation would in case of $N$ domains require $N \times (N - 1) / 2 $ generators. Since we consider $N=3$ here, we have trained a total of 6 generators for CycleGAN and MUNIT. The advantage of SDIT with respect to this non-scalable models would be even more evident for an increased number of domains.

\begin{table}[t]
\setlength{\tabcolsep}{1.5pt}
\resizebox{\columnwidth}{!}{
\begin{tabular}{c|cccc|c|c}

\hline
Method &Yellow & Blue & Green & Orange & Mean & Num E/G \\
\hline
CycleGAN  &93.4/0.599    &95.1/0.601   &93.4/0.584  &92.3/0.587    & 93.5/0.592 & 6/6    \\
IcGAN  &92.2/0.581    &93.5/0.592   &92.8/0.579  &92.1/0.589    & 92.6/0.585 & 1/1    \\
StarGAN  &95.9/0.591    &95.3/0.602   &96.0/0.590  &94.2/0.584    & 95.3/0.591 & 1/1    \\
MUNIT  &97.3/0.607    &\textbf{97.1}/0.603   &\textbf{97.2}/0.599 &96.8/0.621 &\textbf{97.2}/0.608   & 6/6    \\
\hline
SDIT  &\textbf{97.6}/\textbf{0.610}    &96.6/\textbf{0.607}   &\textbf{97.3}/\textbf{0.604}  &97.1/\textbf{0.627}    &\textbf{97.1}/\textbf{0.612}    & 1/1      \\
\hline
\hline
Real  image  &98.5/0.652   &98.6/0.652   &97.8/0.653  &98.8/0.652   &98.4/0.652   & -/-       \\
\hline
\end{tabular}
}
  \caption{\small Reverse classification accuracy ($ \% $) and LPIPS on the color dataset. For both metrics, the higher the better.}
  \vspace{-8mm}
  \label{table:color}%
\end{table}

\subsection{Scene translation}\label{sec:scene}
Finally, we train our model on the photo and artworks dataset~\cite{zhu2017unpaired}.  
Differently from the model used for faces and color objects, here we consider the variant of our model without attention.
This difference is due to the fact that previous datasets had a foreground that needed to be changed (object) and a fixed background, whereas in the scene case we need the generator to learn a global image translation instead of a local one, and thus background must also be changed.

Figure~\ref{fig:color_art}(b) shows several representative examples of the different methods. The conclusions are similar to previous experiments: SDIT maps the input (\textit{photo}) to other domains with diversity while using a single model. Table~\ref{table:arts} also confirms this, showing how the proposed method achieves excellent scores with only one scalable model.

\begin{table}[t]
\setlength{\tabcolsep}{1.5pt}
\resizebox{\columnwidth}{!}{
\begin{tabular}{c|cccc|c|c}

\hline
Method &Photo & Cezanne & Ukiyoe & Monet & Mean & Num E/G \\
\hline
CycleGAN  &52.8/0.684    &57.4/0.654   &56.1/0.674  &60.9/0.648    & 56.8/0.665 & 6/6     \\
IcGAN  &50.9/0.697    &56.8/0.663   &55.1/0.677  &59.7/0.651    & 55.6/0.671 & 1/1     \\
StarGAN  &60.1/0.694    &61.5/0.667   &61.3/0.689  &62.7/0.663    & 61.3/0.678 & 1/1     \\
MUNIT  &\textbf{66.2}/0.763    &\textbf{67.9}/0.784   &\textbf{67.2}/0.791 &63.9/0.778    &\textbf{66.3}/0.779   & 6/6    \\
\hline
SDIT  &65.6/\textbf{0.816}    &63.4/\textbf{0.806}   &65.3/\textbf{0.829}  &\textbf{66.4}/\textbf{0.802}    &65.1/\textbf{0.828} & 1/1       \\
\hline
\hline
Real  image  &70.2/0.856   &72.4/0.874   &69.9/0.884  &71.7/0.864   &71.1/0.869 & -/-     \\
\hline
\end{tabular}
}
  \caption{\small Reverse classification accuracy ($ \% $) and LPIPS on the artworks dataset. For both metrics, the higher the better.}
  \vspace{-8mm}
  \label{table:arts}%
\end{table}

\section{Conclusion}
We have introduced SDIT to perform image-to-image translation with scalability and diversity using a simple and compact network. The key challenge lies in  controlling the two functions separately without conflict. 
We achieve scalability by conditioning the encoder with the target domain label, and diversity by applying conditional instance normalization in the bottleneck. 
In addition, the use of attention on the latent represent further improves the performance of image translation, allowing the model to mainly focus on domain-specific areas instead of the unrelated ones. The model has limited applicability for domains with large variations (for example, faces and paintings in a single model) and works better when the domains have characteristics in common.

\minisection{Acknowledgements.}
Y. Wang acknowledges the Chinese Scholarship Council (CSC) grant
No.201507040048. L. Herranz acknowledges the
European Union research and innovation program under the Marie
Skłodowska-Curie grant agreement No. 6655919. This work was supported
by TIN2016-79717-R, and the CHISTERA project M2CR
(PCIN-2015-251) of the Spanish Ministry, the CERCA
Program of the \emph{Generalitat de Catalunya}. We also acknowledge the generous GPU
support from NVIDIA.
\bibliographystyle{ACM-Reference-Format}
\bibliography{longstrings,refs}

\end{document}